\documentclass[journal]{IEEEtran}
\usepackage{times}
\usepackage{epsfig}
\usepackage{graphicx}
\usepackage{amsmath}
\usepackage{amssymb}
\usepackage{float}
\usepackage{booktabs}
\usepackage{xcolor}

\newcommand{\hou}{}

\ifCLASSINFOpdf
\else
\fi
\begin{document}
%
\title{MW-GAN: Multi-Warping GAN for Caricature Generation with Multi-Style Geometric Exaggeration}
%
%
%

\author{Haodi~Hou,
        Jing~Huo,
        Jing~Wu,
        Yu-Kun~Lai,
        and~Yang~Gao,~\IEEEmembership{Member,~IEEE}%
        \thanks{This work is supported by Science and Technology Innovation 2030 New Generation Artificial Intelligence Major Project (2018AAA0100905), the National Natural Science Foundation of China (61806092), Natural Science Foundation of Jiangsu Province (BK20180326), the Fundamental Research Funds for the Central Universities (02021438008) and the Collaborative Innovation Center of Novel Software Technology and Industrialization. \textit{(Corresponding author: Jing Huo)}}
        \thanks{H. Hou, J. Huo and Y. Gao are with the State Key Laboratory for Novel Software Technology, Nanjing University, No. 163 Xianlin Avenue, Nanjing, 210023, Jiangsu, China. E-mail: hhd@smail.nju.edu.cn, huojing@nju.edu.cn, gaoy@nju.edu.cn.}%
        \thanks{J. Wu and Y.-K. Lai are with the School of Computer Science \& Informatics, Cardiff University, UK. E-mail: \{WuJ11,LaiY4\}@cardiff.ac.uk.}}

%
%

\markboth{IEEE Transactions on Image Processing}%
{Shell \MakeLowercase{\textit{et al.}}: Bare Demo of IEEEtran.cls for IEEE Journals}
%



\maketitle

\begin{abstract}
Given an input face photo, the goal of caricature generation is to produce stylized, exaggerated caricatures that share the same identity as the photo. It requires simultaneous style transfer and shape exaggeration with rich diversity, and meanwhile preserving the identity of the input. To address this challenging problem, we propose a novel framework called Multi-Warping GAN (MW-GAN), including a style network and a geometric network that are designed to conduct style transfer and geometric exaggeration respectively. 
We bridge the gap between the style/landmark space and their corresponding latent code spaces 
by a dual way design, so as to generate caricatures with arbitrary styles and geometric exaggeration, which can be specified either through random sampling of latent code or from a given caricature sample. Besides, we apply identity preserving loss to both image space and landmark space, leading to a great improvement in quality of generated caricatures. Experiments show that caricatures generated by MW-GAN have better quality than existing methods.
\end{abstract}


%
\IEEEpeerreviewmaketitle
\begin{figure*}[t]
  \centering
  \includegraphics[width=1\textwidth]{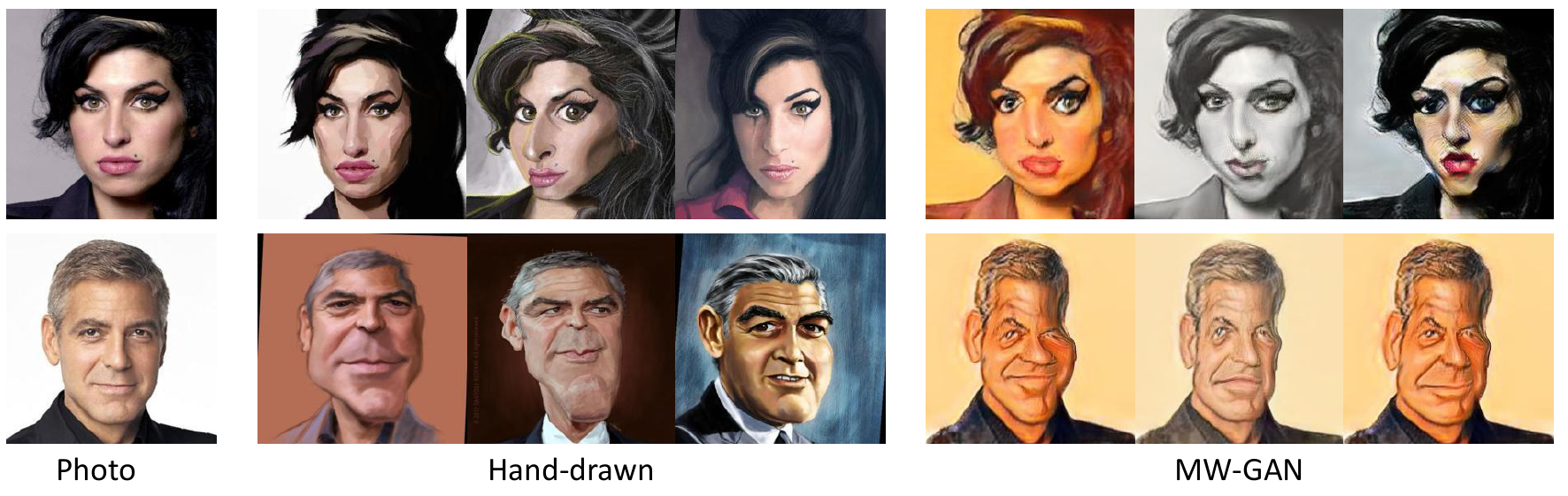}
  \caption{Caricature diversity. The first column shows input photos. The following three columns are caricatures drawn by artists. Caricatures in the last three columns are generated by our MW-GAN with photos in the first column as input. It shows that artists can draw caricatures with various texture styles and exaggerations, and our MW-GAN is designed to model these diversities.}
  \label{intro_figure}
\end{figure*}

\section{Introduction}\label{sec:introduction}
%
%
%
%
\IEEEPARstart{C}{aricatures} are artistic drawings of faces with exaggeration of facial features to emphasize the impressions of or intentions towards the subject. As an art form, caricatures have various depiction styles, such as sketching, pencil strokes and oil painting, and various exaggeration styles to express different impressions and emphasize different aspects of the subject. Artists have their own subjectivity and different skills which also contribute to the diversity of caricatures. As shown in Figure~\ref{intro_figure}, caricatures drawn by artists can have various texture styles and different shape exaggerations even for the same subject. These varieties in caricature generation make caricatures a fascinating art form with long-lasting popularity.

However, such diversity has not been achieved in computerized generation of caricatures. Early works generate caricatures through amplifying the difference from the mean face~\cite{816567,Le2011,liao2004automatic} or automatically learning rules from paired photos and caricatures. However, these methods can only generate caricatures with a specific style. The recent style transfer methods~\cite{Gatys,style_speed1,style_speed2,AdaIN} and image translation methods~\cite{CoGAN,unit,cycleGAN,MUNIT,DTN,CDAAE} based on Convolutional Neural Networks (CNNs) and Generative Adversarial Nets (GANs)~\cite{GAN} have achieved appealing results on image style translation in texture and color. However, these methods are not designed to deal with geometric shape exaggeration in caricatures. The recent GAN-based caricature generation methods~\cite{WenbinLi,warpGAN,cariGANs} can generate caricatures with reasonable exaggerations, but still \hou{lack variety in geometric exaggeration}, leaving a gap between computer generated and real caricatures. \hou{CariGAN by Li et al.~\cite{WenbinLi} translates both texture and shape in a single network, and treats the translation as a deterministic mapping function, which restricts the diversity of the generated caricatures. 
WarpGAN by Shi et al.~\cite{warpGAN} and CariGANs by Cao et al.~\cite{cariGANs} separately render the images' texture and exaggerate face shapes. Though they can generate caricatures with appealing texture styles and meaningful exaggerations, their exaggeration is fixed according to the input photo.}

To tackle this issue, in this paper, we propose Multi-Warping GAN for generating caricatures from face photos with a focus on generating various geometric exaggerations. It is a GAN-based framework to generate caricatures with multiple exaggerations by applying Multiple Warping styles to face images, and is thus called Multi-Warping GAN (MW-GAN). \hou{To allow for the diversity of both texture and geometric exaggeration styles, MW-GAN is designed to have a style network and a geometric network.}
The style network is trained to render images with different texture and coloring styles, while the geometric network learns the exaggeration in the landmark space and warps images accordingly. In both networks, we propose to use latent codes to control the texture and exaggeration styles respectively. The diversity is achieved by random sampling of the latent codes or extracting them from sample caricatures. To correlate the latent codes with meaningful texture styles and shape exaggerations, we propose a dual way architecture, which simultaneously translates photos into caricatures and caricatures into photos, with the aim to provide more supervision on the latent code. With the dual way design, cycle consistency loss on latent code can be introduced. This allows us to not only get more meaningful latent codes, but also obtain better generation results compared with using the single way design. \hou{Besides, compared with \cite{warpGAN} and \cite{cariGANs}, our method supports multiple exaggeration styles for the same input photo.}

In addition to diversity, another challenge is the identity preservation in generated caricatures. \hou{Previously, Shi et al.~\cite{warpGAN} have proposed to use an identity preservation loss which is defined in the image space to preserve identity.} Observing that caricaturization involves both style translation and shape deformation, to preserve the identity of the subject in the input photo, we deploy identity recognition loss in both image space and landmark space when training the networks. 
The loss in both spaces leads to remarkable quality improvement of the generated caricatures.

We conducted ablation studies to verify the effectiveness of the dual way architecture in comparison with the single way design, and the introduction of the landmark constraints in the identity recognition loss. We compared 
our method with the state-of-the-art caricature generation methods in terms of the quality of the generated caricatures. And we demonstrated 
the diversity of both the texture and exaggeration styles in the generated caricatures using our method. Results showed 
both the effectiveness of our method and its superiority over the state-of-the-arts.

In summary, the contributions of our work are as follows:

\begin{enumerate}
\item Our method is the first to focus on the diversity of geometric exaggeration 
in caricature generation, and we propose a GAN-based framework that can generate caricatures with arbitrary texture and exaggeration styles.

\item 
Our framework proposes a dual way design to learn more meaningful relations between the image style/shape exaggeration spaces and their corresponding latent code spaces, and enables the specification of the styles and exaggerations of generated caricatures from caricature samples.

\item To preserve the identity of the subject in the photo, we also deploy identity recognition loss in both image space and landmark space when training the network, which leads to remarkable improvement in the quality of generated caricatures.
\end{enumerate}

We compare our results with those from the state-of-the-art methods, and demonstrate the superiority of our method in terms of both quality and diversity of the generated caricatures.

 

\section{Related work}

\subsection{Style Transfer}
Since CNNs have achieved great success in understanding the semantics in images, it is widely studied to apply CNNs to style transfer. The ground-breaking work of Gatys et al.~\cite{Gatys} presented a general neural style transfer method that can transfer the texture style from a style image to a content image. Following this work, many improved methods~\cite{style_speed1,style_speed2} have been proposed to speed up the transfer process by learning a specific style with a feed-forward network and transfer an arbitrary style in real time through adaptive instance normalization~\cite{AdaIN}. Despite the achievements in transferring images with realistic artistic styles, these methods can only change the texture rendering of images, but are not designed to make the geometric exaggeration required in caricature generation. In our MW-GAN, a style network together with a geometric network are used to simultaneously render the image's texture style and exaggerate its geometric shape, with the aim to generate caricatures with both realistic texture styles and meaningful shape exaggerations.

\subsection{Generative Models for Image Translation}
The success of Generative Adversarial Nets (GANs)~\cite{GAN} has inspired a series of work on cross-domain image translation~\cite{pix2pix,cycleGAN,tang2020unified}. The pix2pix network~\cite{pix2pix} is trained with a conditional GAN, and needs supervision from paired images which are hard to get. Triangle GAN~\cite{triangleGAN} achieved semi-supervised image translation by combining a conditional GAN and a Bidirectional GAN~\cite{bidirectionalGAN} with a triangle framework. There have been efforts to achieve image translation in a totally unsupervised manner through shared weights and latent space~\cite{unit,CoGAN}, cycle consistency~\cite{cycleGAN}, and making use of semantic features~\cite{DTN}. 
The above methods treat image translation as a one-to-one mapping. Recently more methods have been proposed to deal with image translation with multiple styles. Augmented Cycle GAN~\cite{augmentedCycleGAN} extends cycle GAN to multiple translations by adding a style code to model various styles. MUNIT~\cite{MUNIT} and CDAAE~\cite{CDAAE} disentangle an image into a content code and a style code, so that a single input image can be translated to various output images by sampling different style codes. These methods can successfully translate images between different domains, and can render with various texture styles in one translation. However, these translations mostly keep the image's geometric shapes unchanged, which is not suitable for caricature generation. By contrast, we separately model the two aspects, texture rendering and geometric exaggeration, and achieve both translations in a multiple style manner. That is, our model can generate caricatures with various texture styles and diverse geometric exaggerations for a given input.

\begin{figure*}[t]
  \centering
  \includegraphics[width=0.97\textwidth]{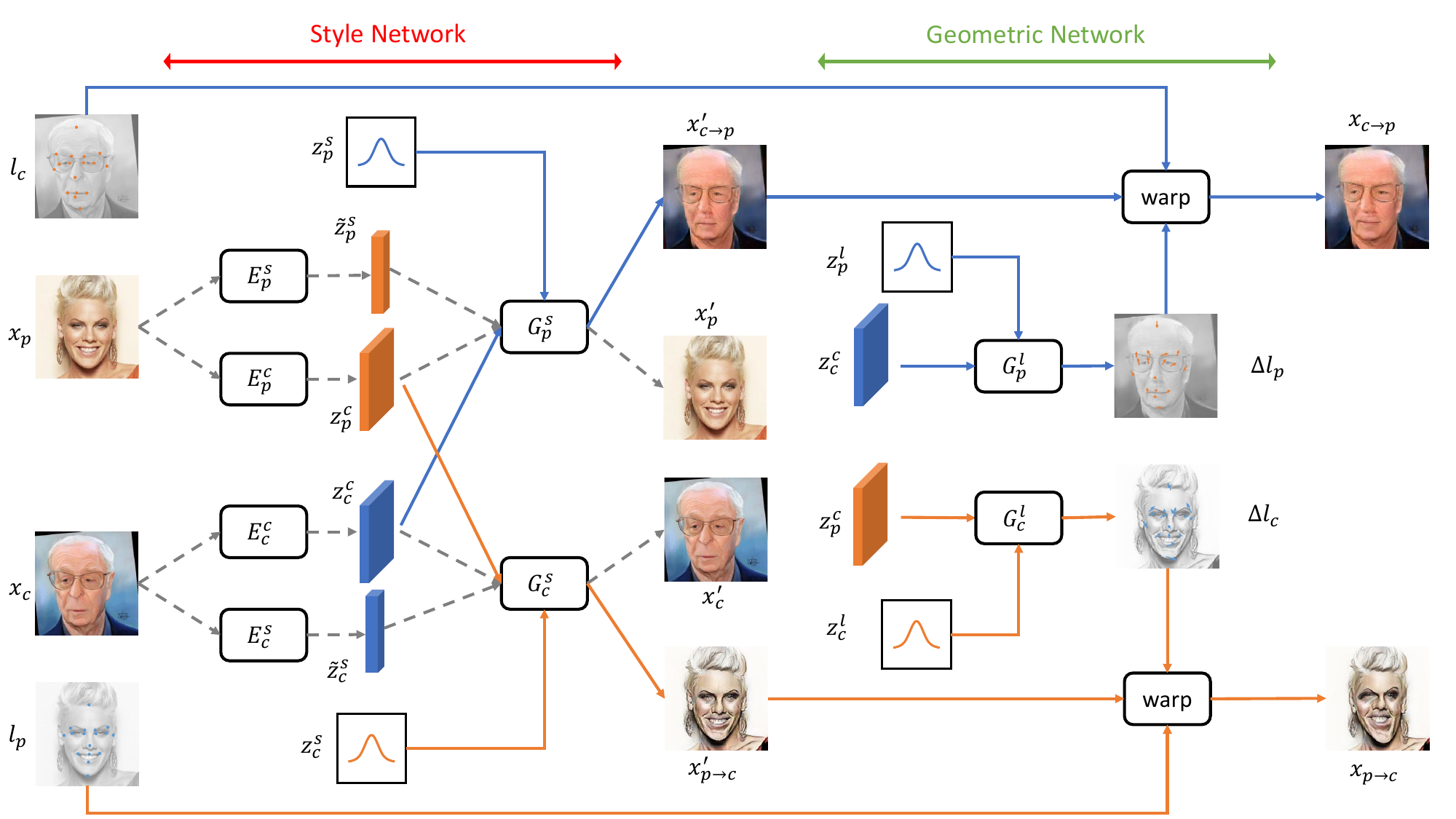}
  \caption{The network architecture of the proposed multi-warping GAN. The left part is the style network and the right part is the geometric network. The gray dashed arrows denote the flow of two auto-encoders, with the upper one being the auto-encoder of photos and the lower one for caricature reconstruction. The orange arrows denote the flow of photo-to-caricature generation and the blue arrows for caricature-to-photo generation. In our dual way framework, we assume the caricature and the photo share the same content feature space but have separate style spaces. $E_p^c$ and $E_c^c$ are two encoders to encode the content of photos and caricatures respectively. Similarly defined, $E_p^s$ and $E_c^s$ are two encoders to encode the style of photos and caricatures. Gaussian distribution is imposed on their outputs $\tilde{z}_p^s$ and $\tilde{z}_c^s$, so that we can sample style codes from Gaussian when translating photos to caricatures or vice versa. As for the geometric exaggeration, we assume that it depends on the content latent code ($z_c^c$ or $z_p^c$) and a landmark transformation latent code ($z_c^l$ or $z_p^l$), where the former captures the characteristics of the input face, while the latter represents the artistic style. That is to say, a generator network $G_c^l$ ($G_p^l$) is trained to output a landmark displacement map $\Delta l_c$ ($\Delta l_p$) with $z_p^c$ ($z_c^c$) and $z_c^l$ ($z_p^l$) as input. To get the final translated caricature $x_{p\rightarrow c}$ (translated photo $x_{c\rightarrow p}$), we conduct geometric exaggeration on the stylized image $x'_{p\rightarrow c}$ ($x'_{c\rightarrow p}$) through warping according to its original facial landmarks $l_p$ ($l_c$) and the learned landmark displacements $\Delta l_c$ ($\Delta l_p$). Here we only show the image translation flow of the geometric network, and more details are illustrated in Section~\ref{section_geonet}.
  }
  \label{ourmodel}
\end{figure*}

\subsection{Caricature Generation}
Caricature generation has been studied for a long time. Traditional methods translate photos to caricatures using computer graphics techniques. The first interactive caricature generator was presented by Brennan et al.~\cite{brennan1985caricature}. The caricature generator allows users to manipulate photos interactively to create caricatures. Following their work, rule-based methods were proposed~\cite{816567,liao2004automatic,Le2011} to automatically amplify the difference from the mean face. Example-based methods~\cite{liang2002,Shet2005} can automatically learn rules from photo-caricature pairs. Although these methods can generate caricature automatically or semi-automatically, they suffer from some limitations, such as the need of human interactive manipulation and paired data collection. Moreover, caricatures generated by these early methods are often unrealistic and lack diversity.

Since GANs have made great progress in image generation, many GAN-based methods for caricature generation were presented recently. Some of these methods translate photos to caricatures with a straightforward network~\cite{DTN,WenbinLi}, while others translate the texture style and geometric shapes separately~\cite{warpGAN,cariGANs}. For the straightforward methods, \hou{Domain Transfer Network (DTN)~\cite{DTN}} uses a pretrained neural network to extract semantic features from input so that semantic content can be preserved during translation. CariGAN by Li et al.~\cite{WenbinLi} adopts facial landmarks as an additional condition to enforce reasonable exaggeration and facial deformation. As these methods translate both texture and shape in a single network, it is hard for them to achieve meaningful deformation or to balance identity preservation and shape exaggeration. By contrast, WarpGAN~\cite{warpGAN} and CariGANs by Cao et al.~\cite{cariGANs} separately render the image's texture and exaggerates its shape. Although they can generate caricatures with realistic texture styles and meaningful exaggerations, WarpGAN and CariGANs~\cite{cariGANs} still suffer from lacking exaggeration variety. \hou{Specifically, when the input is specified, they can only generate caricatures with a fixed exaggeration.} However, in real world, it is common that different artists draw caricatures with different exaggeration styles for the same photo. In this paper, we design a framework that is able to model the variety of both texture styles and geometric exaggerations and propose the first model that can generate caricatures with diverse styles in both texture and exaggeration for one input photo.

\section{Multi-Warping GAN}
\label{Sec_MWGAN}
In this section, we describe the network architecture of the proposed Multi-Warping GAN and the loss functions used for training.

\subsection{Notations}
Let $x_p\in \mathcal{X}_p$ denote an image in the photo domain $\mathcal{X}_p$, and $x_c\in \mathcal{X}_c$ denote an image in the caricature domain $\mathcal{X}_c$. Given an input face photo $x_p\in \mathcal{X}_p$, the goal is to generate a caricature image in the space $\mathcal{X}_c$, while sharing the same identity as $x_p$. This process involves two types of transition, texture style transfer and geometric shape exaggeration. Previous works~\cite{cariGANs,warpGAN} can only generate caricatures with a fixed geometric exaggeration style when an input is given. In this paper, we focus on the problem of caricature generation with multiple geometric exaggeration styles, and propose the first framework to deal with it.

The notations used in this paper are as follows. We use $x, z, l, y$ to denote image sample, latent code, landmark and identity label respectively. Subscripts $p$ and $c$ refer to photo and caricature respectively, while superscripts $s$ and $c$ represent style and content. Encoders, generators (a.k.a. decoders) and discriminators are represented by capital letters $E$, $G$ and $D$, respectively.

\subsection{Multi-Warping GAN}
\label{SubSec_MWGAN}
The network architecture of Multi-Warping GAN is shown in Figure~\ref{ourmodel}. It consists of a style network and a geometric network. The style network is designed to render images with different texture and texture styles, while the geometric network aims to exaggerate the face shapes in the input images. The style network works in the image space, while the geometric network is built on landmarks and exaggerates geometric shapes through warping. Both style and geometric networks are designed in a dual way, i.e., there is one way to translate photos to caricatures and also the other way to translate caricatures to photos. \hou{
In this paper, we are mainly interested in translating photos to caricatures. Although it can also be achieved with a single way network, we claim that the dual way design is essential for high-quality generation. For the style network, using a single way is also reasonable and may achieve competitive results. However, using a dual way design has its superiority in constraining the content of the generated caricature, as the dual way model can encode the content of the generated caricature backward and constrain it with the cycle loss. As for the geometric network, using an additional encoder to map the generated caricature back to the landmark latent code is necessary to enforce the network to learn a bidirectional mapping, while a single way model can easily ignore the landmark information.}
We experimentally verified that the dual way framework is more effective compared with the single way design.

In our dual way design, the style and the shape exaggeration are represented by latent codes $z^s$ and $z^l$ respectively. Both latent codes can be sampled from Gaussian distribution or extracted from sample caricature images to achieve the diversity in both style and exaggeration. To train this network, we design a set of loss functions to tighten the corresponding relations between the latent code space and the image space, and to keep identity consistency. In the following, we will explain the details of our style network and geometric network along with the loss functions accordingly.

\subsubsection{Style Network}
During the texture style transfer, the face shape in the image should be preserved. We thus assume that there is a joint shape space, referred to as ``content'' space, shared by both photos and caricatures, while their style spaces are independent. Following MUNIT~\cite{MUNIT}, the style network is composed of two autoencoders for content and style respectively, and is trained to satisfy the constraints in both the image reconstruction process and the style translation process.

The image reconstruction process is shown in Figure~\ref{ourmodel} with gray dashed arrows, and can be formulated as follows:
\begin{equation}
\begin{split}
x'_p=G_p^s(E_p^c(x_p),E_p^s(x_p)),\\
x'_c=G_c^s(E_c^c(x_c),E_c^s(x_c)).
\end{split}
\end{equation}
where $E_p^c$ and $E_p^s$ are content and style encoders for photos. Similarly, $E_c^c$ and $E_c^s$ are content and style encoders  for caricatures. $G_p^s$ and $G_c^s$ are two decoders for photos and caricatures respectively. The image reconstruction loss is defined as the $\ell_1$ difference between the input and reconstructed images:
\begin{equation}
\label{Eq_style_image_recons}
L_{rec\_x}=\Vert x'_p-x_p\Vert_1+\Vert x'_c-x_c\Vert_1.
\end{equation}

The style translation process is shown in Figure~\ref{ourmodel} with coloured arrows, and can be formulated as:
\begin{equation}
\label{Eq_Style_Trans}
\begin{split}
x'_{p\rightarrow c}=G^s_c(E_p^c(x_p), z_c^s),\\
x'_{c\rightarrow p}=G^s_p(E_c^c(x_c), z_p^s),
\end{split}
\end{equation}
where $z_c^s$ and $z_p^s$ are style codes which can be sampled from Gaussian distributions for the two modalities. $x'_{p\rightarrow c}$ is a generated image with the content from the input photo and a caricature style, while $x'_{c\rightarrow p}$ is a generated image with the content from the input caricature and a photo style.

The constraints in the style transfer process are based on three aspects. Firstly, after transfer, the style code of the transferred image should be consistent with the style code of input, i.e.,
\begin{equation}
\label{Eq_style_stylecode_recons}
L_{rec\_s}=\Vert z_p^s-E_p^s(x'_{c\rightarrow p})\Vert_1+\Vert z_c^s-E_c^s(x'_{p\rightarrow c})\Vert_1
\end{equation}
where $x'_{c\rightarrow p}$ and $x'_{p\rightarrow c}$ are defined in Eq.~(\ref{Eq_Style_Trans}).
Secondly, the image content should keep unchanged during the transfer. A cycle consistency loss on the content codes of the input and the transferred images is used, as shown below:
\begin{equation}
\label{Eq_style_cyc_content}
\begin{split}
  L_{cyc\_c}=&\Vert E_p^c(x_p)-E_c^c(x'_{p\rightarrow c})\Vert_1 \\
  +&\Vert E_c^c(x_c)-E_p^c(x'_{c\rightarrow p})\Vert_1
\end{split}
\end{equation}
Thirdly, the transferred image should be able to convert back when passing through the same encoder-decoder and using the original style code. Again a cycle consistency loss on the input and transferred images is used for this constraint:
\begin{equation}
\label{Eq_style_cyc_image}
\begin{split}
L_{cyc\_x}=&\Vert x_p-G_p^s(E_c^c(x'_{p\rightarrow c}), E_p^s(x_p))\Vert_1 \\
+&\Vert x_c-G_c^s(E_p^c(x'_{c\rightarrow p}), E_c^s(x_c))\Vert_1.
\end{split}
\end{equation}
Please note, the second terms in Eq.~(\ref{Eq_style_cyc_content}) and Eq.~(\ref{Eq_style_cyc_image}) constrain the style transfer from caricatures to photos. It is only possible to impose this cycle consistency in our dual way design. It is expected that the cycle consistency can help build the relation between the latent code space and the image space. A single way network from photos to caricatures only is also implemented as a baseline (see Section~\ref{baseline_method}) which is trained without these two terms. Experimental results, as in Section~\ref{sec:experiments}, demonstrate the superior generation results using the dual way design with the cycle consistency loss.

Eqs.~(\ref{Eq_style_image_recons}), (\ref{Eq_style_stylecode_recons}), (\ref{Eq_style_cyc_content}) and
(\ref{Eq_style_cyc_image}) give all the loss functions to train the style network. With the above network architecture, we can generate caricatures with various texture styles by sampling different style codes.

\begin{figure*}[t]
  \centering
  \includegraphics[width=0.98\textwidth]{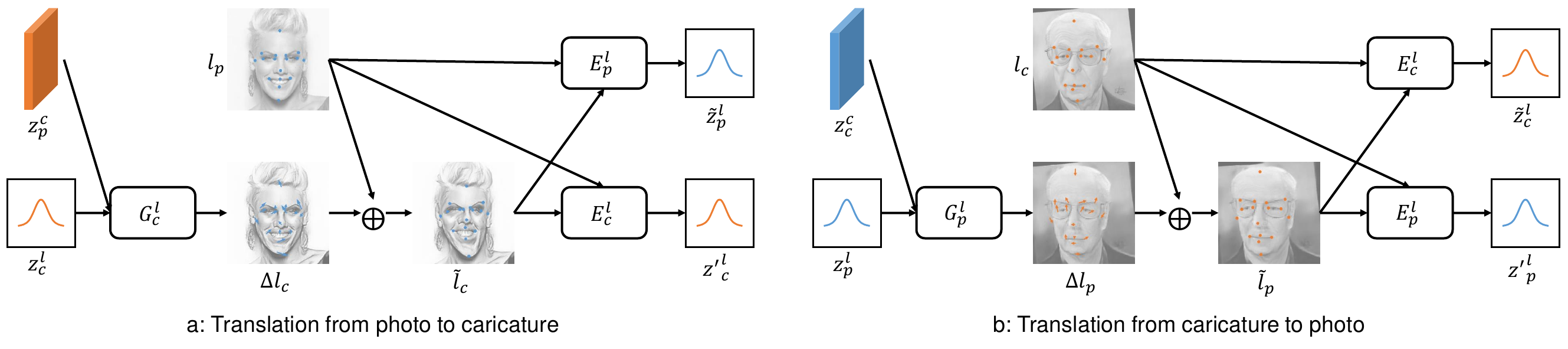}
  \caption{Geometric Network. The left part is the network for learning a transformation from a photo's landmarks to a caricature's landmarks. The right part is the network for the reverse transformation. For the left network, a generator $G_{c}^l$ with the content code of a photo $z_p^c$ and a landmark transformation latent code $z_{c}^l$ (which can be randomly sampled from a Gaussian distribution) as input will output landmark displacement vectors $\Delta l_{c}$. By adding the displacement vectors to the photo's landmark positions, we get the transformed caricature landmarks $\tilde{l}_{c}$. To make the randomly sampled $z_{c}^l$ correlate to meaningful shape transformation styles, we introduce two encoders and force cycle consistency loss on the encoded latent code and sampled latent code. For example, $z_{c}^{'l}=E_c^l(\tilde{l}_{c}, l_p)$ is the encoded latent code, and we force $z_{c}^{'l}$ to be as close as possible to $z_{c}^{l}$.}
  \label{warp}
\end{figure*}

\subsubsection{Geometric Network}
\label{section_geonet}
Geometric exaggeration is an essential feature of caricatures. There are two aspects to consider when modeling caricature exaggerations. One is that exaggerations usually emphasize the subject's characteristics. The other is that they also reflect the skills and preference of the artists. Therefore, in caricature generation, it is natural to model geometric exaggeration based on both the input photo and an independent exaggeration style. By changing the exaggeration style, we can mimic different artists to generate caricatures that have different shape exaggerations for a given input photo. It is straightforward to use a random latent code to model this independent exaggeration style. However, this idea suffers from some problems: 
1) a random latent code may be ignored while training the model to generate realistic caricatures and 2) it is hard to ensure that a random latent code leads to meaningful exaggeration. Thus, we design our geometric network to learn a bidirectional mapping between an exaggeration latent code space and the face landmarks.

In our design, we build a latent code space on landmarks (landmark transformation latent code space) to represent the different exaggeration styles from artists, and use the content code to represent the input photo. 
The geometric network is designed and trained to learn mappings of two directions. The first one is the mapping from a combination of a code in the landmark transformation latent space and a content code to a landmark displacement map, which defines the geometric deformation between the input photo and the caricature to be generated. The landmark displacement map thus captures both the input subject's characteristics and a specific geometric exaggeration style. In the second mapping, a pair of landmarks of photo and caricature is mapped back to the landmark transformation latent code space, so that points in the landmark transformation latent code space are associated with meaningful exaggeration styles. \hou{Geometric exaggeration is finally achieved by warping~\cite{warping} the input photo according to the learned landmark displacements. More concretely, the stylized input image is warped by deforming the image according to its original landmarks and the learned landmark displacements.} In our paper, landmarks from the WebCaricature dataset are used. In real-world applications, face landmarks can be detected with existing detectors.

Following the above assumption, the design of our geometric network is as shown in Figure~\ref{warp}. It consists of one generator ($G_{c}^l$ or $G_{p}^l$) in each way of translation and two encoders ($E_p^l$, $E_c^l$) whose functions will be explained later. Taking the translation from photos to caricatures as an example (shown in Figure~\ref{warp}a), the generator $G_{c}^l$ takes the content code $z_p^c$ and the landmark transformation latent code $z_{c}^l$ as input, and outputs the landmark displacements $\Delta l_{c}$ which are then added to the input landmarks $l_p$ to get the target caricature landmarks $\tilde{l}_{c}$.

The landmark transformation latent code $z_{c}^l$ encodes a shape exaggeration style. To make it correlate to meaningful shape exaggerations, we follow the idea of Augmented CycleGAN~\cite{augmentedCycleGAN} and introduce two encoders $E_p^l$, $E_c^l$ into the geometric network. Again, taking the translation from photos to caricatures for example, the encoder $E_c^l$ takes the landmarks of the photo $l_p$ and the landmarks of the corresponding caricature $\tilde{l}_{c}$ as input, extracts the difference between them and reconstructs the landmark latent code ${z'}_{c}^l$. By introducing the encoder, it enables us 1) to enforce cycle consistency between the randomly sampled latent code $z_{c}^l$ and the encoded latent code ${z'}_{c}^l$ to correlate $z_{c}^l$ to meaningful exaggerations; and 2) to extract $z_{c}^l$ from example caricatures to perform sample guided shape exaggeration. The same applies for the encoder $E_p^l$ used in translation from caricatures to photos (Figure~\ref{warp}b).

Basically, to train the geometric network, we have the landmark transformation latent code reconstruction loss:
\begin{equation}
\label{Eq_Cycle_LandmarkLatentCode}
L_{rec\_z\_l}=\Vert z_{p}^l-E_p^l(\tilde{l}_{p}, l_c)\Vert_1+\Vert z_{c}^l-E_c^l(\tilde{l}_{c}, l_p)\Vert_1
\end{equation}
where the first term is the reconstruction loss of $z_{p}^l$, the landmark transformation latent code from photos to caricatures. The second term is the reconstruction loss of $z_{c}^l$ and is defined in a similar way.

Besides the above loss, we use LSGAN~\cite{lsgan} to match the generated landmarks with real ones:
\begin{align}
  L^G_{gan\_l}=&{\Vert 1-D_p^l(\tilde{l}_{p})\Vert}^2+{\Vert 1-D_c^l(\tilde{l}_{c})\Vert}^2, \label{Eq_LSGAN_Landmark0} \\
  \nonumber L^D_{gan\_l}=&{\Vert 1-D_p^l(l_p)\Vert}^2+{\Vert D_p^l(\tilde{l}_{p})\Vert}^2
\\+&{\Vert 1-D_c^l(l_c)\Vert}^2+{\Vert D_c^l(\tilde{l}_{c})\Vert}^2.\label{Eq_LSGAN_Landmark}
\end{align}
where Eq.~(\ref{Eq_LSGAN_Landmark0}) is the loss for generators and Eq.~(\ref{Eq_LSGAN_Landmark}) is the loss for discriminators. The objective of generators is to make the generated caricature landmarks $\tilde{l}_{c}$ or photo landmarks $\tilde{l}_{p}$ indistinguishable from real landmarks, i.e., the output of the discriminator with the generated landmarks as input becomes 1. On the other hand, the objective of discriminators is to discriminate between the real photo landmarks $l_p$ and the generated photo landmarks $\tilde{l}_{p}$, as well as to discriminate between the real caricature landmarks $l_c$ and the generated caricature landmarks $\tilde{l}_{c}$.

Similar to above loss for generated landmarks, we define the loss for generated images:
\begin{align}
  L^G_{gan\_x}=&{\Vert 1-D_p^x(x_{c\rightarrow p})\Vert}^2+{\Vert 1-D_c^x(x_{p\rightarrow c})\Vert}^2, 
  \label{Eq_LSGAN_Image0}
  \\
  L^D_{gan\_x}=&{\Vert 1-D_p^x(x_p)\Vert}^2+{\Vert D_p^x(x_{c\rightarrow p})\Vert}^2 \nonumber \\
  +&{\Vert 1-D_c^x(x_c)\Vert}^2+{\Vert D_c^x(x_{p\rightarrow c})\Vert}^2.
\end{align}
The definition of the above two losses are in the same way as the losses in Eqs. (\ref{Eq_LSGAN_Landmark0}) and (\ref{Eq_LSGAN_Landmark}), except that the generated landmarks are now changed to images.

We also use LSGAN~\cite{lsgan} to match all the latent codes (including both landmark transformation latent code and the style latent code, except content code) to Gaussian:
\begin{align}
  L^G_{gan\_z}=&{\Vert 1-D^z(\tilde{z})\Vert}^2, \label{Eq_LG_GAN_z} \\
  L^D_{gan\_z}=&{\Vert 1-D^z(z)\Vert}^2+{\Vert D^z(\tilde{z})\Vert}^2. \label{Eq_LD_GAN_z} 
\end{align}
Here, $\tilde{z}$ is latent codes encoded by neural encoders, while $z$ is latent codes sampled from Gaussian distribution. Eq.~(\ref{Eq_LG_GAN_z}) is the loss for the generator and Eq.~(\ref{Eq_LD_GAN_z}) is the loss for the discriminator. The objective of the generator is to make the discriminator unable to tell whether the encoded latent code $\tilde{z}$ is sampled from Gaussian or not. And the discriminator's objective is to try to discriminate between these two kinds of codes.

\subsubsection{Identity Preservation}
Identity preservation in the generated caricatures becomes more challenging with the explicit geometric deformation introduced. As a result, in addition to preserving identity in the image space as in~\cite{warpGAN}, we add further constraints on identity in the landmark space. Two discriminators are added to classify the identity from both the image and the landmarks.
\begin{align}
  L_{id\_x}=&-\log(D_{id}^x(y_p, x_p))-\log(D_{id}^x(y_p, x_{p\rightarrow c}))
  \nonumber
  \\&-\log(D_{id}^x(y_c, x_c))-\log(D_{id}^x(y_c, x_{c\rightarrow p}))\\
  L_{id\_l}=&-\log(D_{id}^l(y_p, l_p))-\log(D_{id}^l(y_p, \tilde{l}_{c}))
 \nonumber 
  \\&-\log(D_{id}^l(y_c, l_c))-\log(D_{id}^x(y_c, \tilde{l}_{p})).
\end{align}
Here, the two discriminators $D_{id}^x$ and $D_{id}^l$ are both classifiers for face identity, except that $D_{id}^x$ takes images as input while $D_{id}^l$ takes landmarks as input. $y_p$ and $y_c$ are the identity labels for the corresponding photos and caricatures. The label only represents the face identity no matter what style it is or whether it is photo or caricature. As for the translated images $x_{p\rightarrow c}$ and $x_{c\rightarrow p}$, $y_p$ and $y_c$ are labels of the corresponding input images $x_p$ and $x_c$ respectively, so that the translated images have the same identity as the input images. It is similar for the translated landmarks $\tilde{l}_c$ and $\tilde{l}_p$. To be clear, these two discriminators are trained with the whole framework end-to-end, without pretraining.

\subsubsection{Overall Loss}
In summary, training the proposed MW-GAN is to minimize the following types of loss functions: 1) the reconstruction loss of the image $L_{rec\_x}$, the style latent code $L_{rec\_s}$, and the landmark transformation latent code $L_{rec\_z\_l}$, and the cycle consistency loss of the content code $L_{cyc\_c}$ and of the image $L_{cyc\_x}$; 2) the generative adversarial loss pairs on images $L_{gan\_x}^G$, $L_{gan\_x}^D$, landmarks $L_{gan\_l}^G$, $L_{gan\_l}^D$, and latent codes $L_{gan\_z}^G$, $L_{gan\_z}^D$; and 3) the identify loss on the image $L_{id\_x}$, and on the landmarks $L_{id\_l}$.

Our framework is trained by optimizing the following overall objective functions on encoders, generators, and discriminators:
\begin{equation}
\label{opt_g}
\begin{split}
  \min_{E,G} &\lambda_1L_{rec\_x}+\lambda_2(L_{rec\_s}+L_{rec\_z\_l}+L_{cyc\_c}+L_{cyc\_x}) \\+ &\lambda_3L_{id\_x}+\lambda_4L_{id\_l}+\lambda_5(L_{gan\_x}^G+L_{gan\_l}^G+L_{gan\_z}^G),
\end{split}
\end{equation}
\begin{equation}
\label{opt_d}
\min_{D}\lambda_3L_{id\_x}+\lambda_4L_{id\_l}+\lambda_5(L_{gan\_x}^D+L_{gan\_l}^D+L_{gan\_z}^D).
\end{equation}
Here, $E$ and $G$ denote the encoder and generator networks, whereas $D$ denotes the discriminator networks. $\lambda_1$, $\lambda_2$, $\lambda_3$, $\lambda_4$ and $\lambda_5$ are weight parameters to balance the influence of the different loss terms in the overall objective function. The whole framework 
is trained
in an end-to-end manner. With a mini batch of training data (including images, landmarks and identity labels), we update parameters of $E$, $G$ and $D$ alternately. That is to say, in one step of iteration, we first optimize all the discriminators with Eq. (\ref{opt_d}) and then optimize all the encoders and generator with Eq. (\ref{opt_g}).

\subsection{Degradation to Single Way Baseline}
\label{baseline_method}
\begin{figure}[t]
  \centering
  \includegraphics[width=0.48\textwidth]{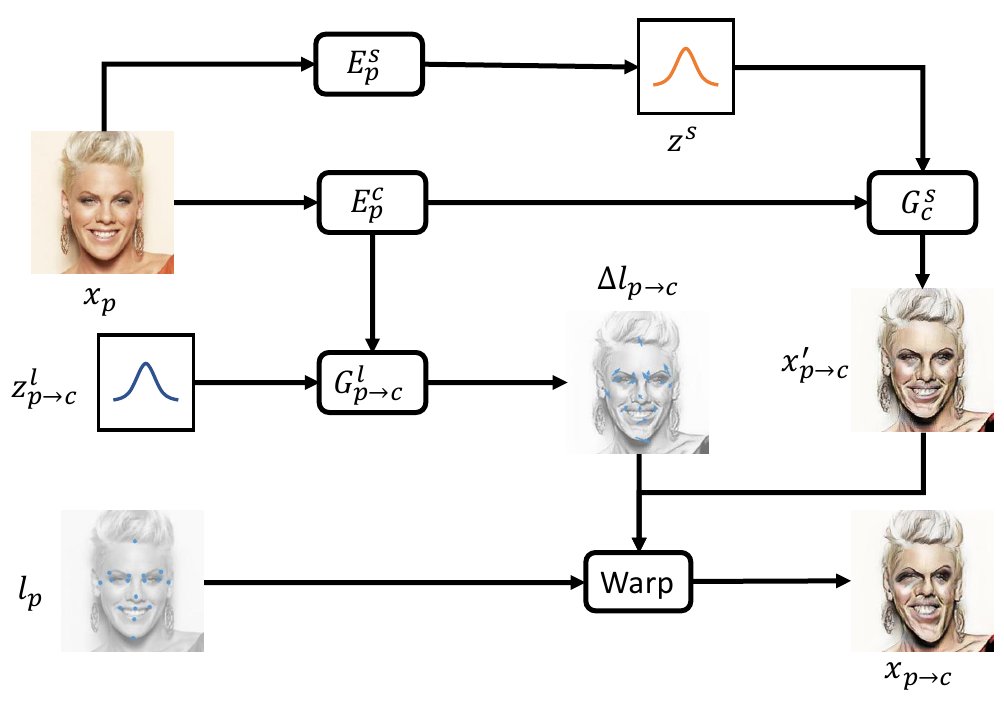}
  \caption{Framework of baseline single way GAN. The upper two rows denote the style network and the lower two rows denote the geometric network.}
  \label{baseline}
\end{figure}

Notice for the caricature generation task, we can also degrade the above framework to a single way network, i.e. only from photos to caricatures as shown in Figure~\ref{baseline}. However, we found that without the cycle consistency loss, the single way network for caricature generation performs not as good as 
the dual way design. Here, we give details of the single way network and this forms a baseline method in our experiments.

As shown in Figure~\ref{baseline}, the degraded single way network also consists of a style network and a geometric network. The style network is used to render an input photo with a caricature style while preserving the geometric shapes. It consists of two encoders $E^s_p$, $E^c_p$ and one generator $G^s_c$. The content encoder $E^c_p$ extracts the feature map $z^c_p$ that contains the geometric information of the input photo, while the style encoder $E^s_p$ extracts texture style $z^s_p$ from the input. The style code $z^s_p$ is adapted to a Gaussian distribution and affects the image's style through adaptive instance normalization~\cite{AdaIN}. The geometric network exaggerates the face in the rendered image by warping it according to the landmark displacements ${\Delta}l_{c}$. To achieve multi-style exaggerations, we assume that ${\Delta}l_{c}$ is controlled by not only the photo's content $z^c_p$ but also a landmark transformation latent code $z^l_{c}$ that follows a Gaussian distribution.

\begin{figure*}[t]
  \centering
  \includegraphics[width=0.95\textwidth]{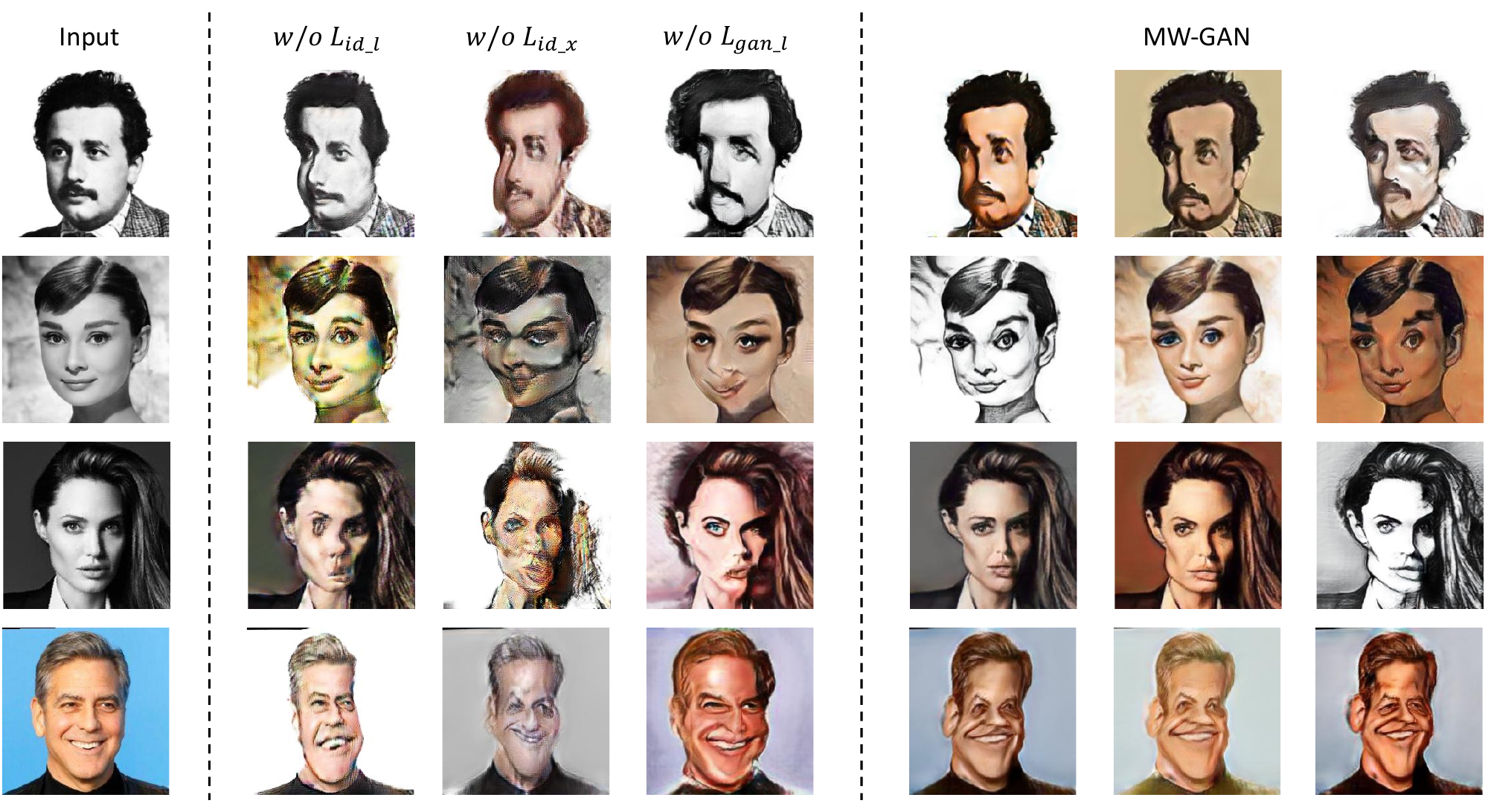}
  \caption{Ablation study. For MW-GAN variants, we generate one caricature for each input. For MW-GAN, we generate three caricatures for each input by sampling different style latent codes and landmark transformation latent codes.}
  \label{ablation}
\end{figure*}

\begin{table*}
  \centering
  \caption{Ablation study. Comparison of the three variants of MW-GAN, the single way baseline and MW-GAN. ``ACC'' is short for the rank-1 accuracy.}
  \label{ablation_table}
  \begin{tabular}{l c c c c c c}
    \toprule
    Method & $w/o$ $L_{id\_l}$ & $w/o$ $L_{id\_x}$ & $w/o$ $L_{gan\_l}$ & baseline & MW-GAN \\
    \midrule
    FID & 47.53 & 56.44 & 41.09 & 57.38 & \textbf{36.29}\\
    ACC & 73.68\% & 37.95\% & 59.49\% & 43.59\% & \textbf{74.87\%}\\
    \bottomrule
  \end{tabular}
\end{table*}

This straightforward framework can increase the variety of exaggeration styles through various landmark transformation latent codes. However, it suffers from some limitations. Firstly, 
as it is a one-way framework without cycle consistency, there lacks supervision to relate the generated caricatures with the landmark transformation latent code. Then with the use of discriminator, the model may ignore the landmark transformation latent code in training which actually restricts the geometric deformation.
Secondly, this one-way framework lacks supervision to bridge the gap between the latent space of $z^l_{c}$ and the landmark space. Thus the learned landmark latent code may have no reference to real landmarks. In experiments, we compare the generation results using our dual way network with using the single way network, and demonstrate the advantages of our dual-way design.

\section{Experiments}
\label{sec:experiments}
We conducted experiments on the WebCaricature dataset~\cite{WebCaricature2,WebCaricature}. We will first describe the details of the dataset and the training process,
and then demonstrate the effectiveness of the dual way design and the landmark based identity loss through ablation studies. We will show the ability of our method to generate caricatures with a variety of both texture and exaggeration styles. And finally, we will compare the caricatures generated using our method with the previous state-of-the-art methods and show the superiority of our method in terms of the generation quality through qualitative and quantitative comparisons, as well as a perceptual study.

\subsection{Experimental Details}
\label{subsec:experiments_details}
\textbf{Dataset preprocessing.} We trained and tested our network on a public dataset WebCaricature~\cite{WebCaricature2,WebCaricature}. There are 6,042 caricatures and 5,974 photographs from 252 persons in this dataset. We first pre-processed all the images by rotating the faces to make the line between eyes horizontal, and cropping the face using a bounding box which covers hair and ears. In detail, an initial box is first created by passing through the centers of ears, the top of head and the chin. Then the bounding box used is the initial box enlarged by a factor of $1.5$. All processed images are resized to $256\times 256$. We randomly split the dataset into a training set of 202 identities (4,804 photos and 4,773 caricatures) and a test set of 50 identities (1,170 photos and 1,269 caricatures). \hou{The generated caricatures have the same resolution as the inputs, which is $256\times 256$.}
All the images presented in this paper are from identities in the test set. The landmarks used in our experiments are the 17 landmarks provided in the WebCaricature dataset~\cite{WebCaricature2,WebCaricature}.

\begin{figure*}[t!]
  \centering
  \includegraphics[width=0.98\textwidth]{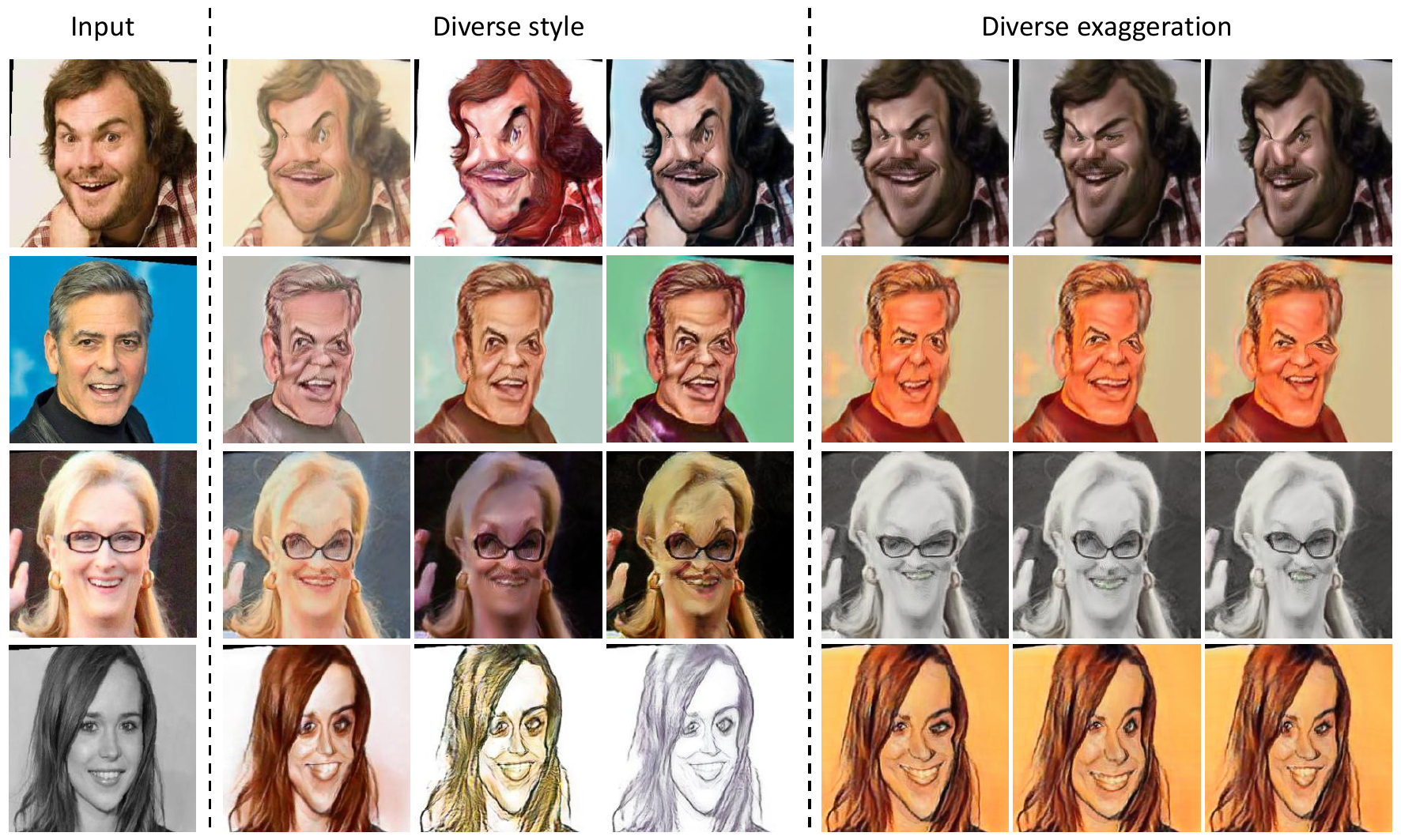}
  \caption{Diversity in texture style and exaggeration of the proposed MW-GAN. The 1st column shows input images. The 2nd-4th columns are corresponding generated caricatures with different texture styles with a fixed exaggeration. The last three columns are generated caricatures with a fixed texture style but different exaggerations.}
  \label{diversity}
\end{figure*}

\textbf{Details of implementation.}
Our framework is implemented with Tensorflow. The style network is modified based on MUNIT~\cite{MUNIT}. We removed the discriminator on the stylized images and added a discriminator on the warped images at the end. We also added a discriminator on the style latent code generated by the style encoder to match it with a Gaussian distribution. In the geometric network, we take $G_{c}^l$ and $E_p^l$ as an example to explain the detailed structure, and $G_{p}^l$ and $E_c^l$ are implemented in the same way. For  $G_{c}^l$, the content code is firstly down-sampled by max pooling with kernel size $3$ and stride $2$, and then is fed into three blocks of $3\times 3$ convolution with stride 1 followed by leaky ReLU with $\alpha=0.01$ and $3\times 3$ max pooling with stride 2. After that, there is a fully connected layer mapping this to a 32-dimensional vector. The landmark latent code is also mapped to a 32-dimensional vector by a fully connected layer. Then the two vectors are concatenated and fed into a fully connected layer to output $\Delta l_{c}$. For $E_p^l$, the two sets of input landmarks (landmarks for photo and landmarks for caricature) are firstly concatenated and then fed into four fully connected layers to give the estimated landmark latent code. All the fully connected layers in $G_{c}^l$ and $E_p^l$ are activated with leaky ReLU, except the last layer. Discriminators for images are composed of 6 blocks of $4\times 4$ convolution with stride 2 and a last layer with full connection, while discriminators for latent codes consist of six layers with full connection. Leaky ReLU is used as activation for all discriminators.

We empirically set $\lambda_1=10$, $\lambda_2=\lambda_5=1.0$, $\lambda_3=0.05$, $\lambda_4=0.01$. We used ADAM optimizers with $\beta_1=0.5$ and $\beta_2=0.999$ to train the whole network. The network is trained for 500,000 steps with batch size of 1. The learning rate is started with 0.0001 and decreased by half every 100,000 steps. The model is trained on a computer with an NVIDIA GeForce RTX 2080 Ti GPU, and the training takes about three days.

\begin{figure}[t]
  \centering
  \includegraphics[width=0.48\textwidth]{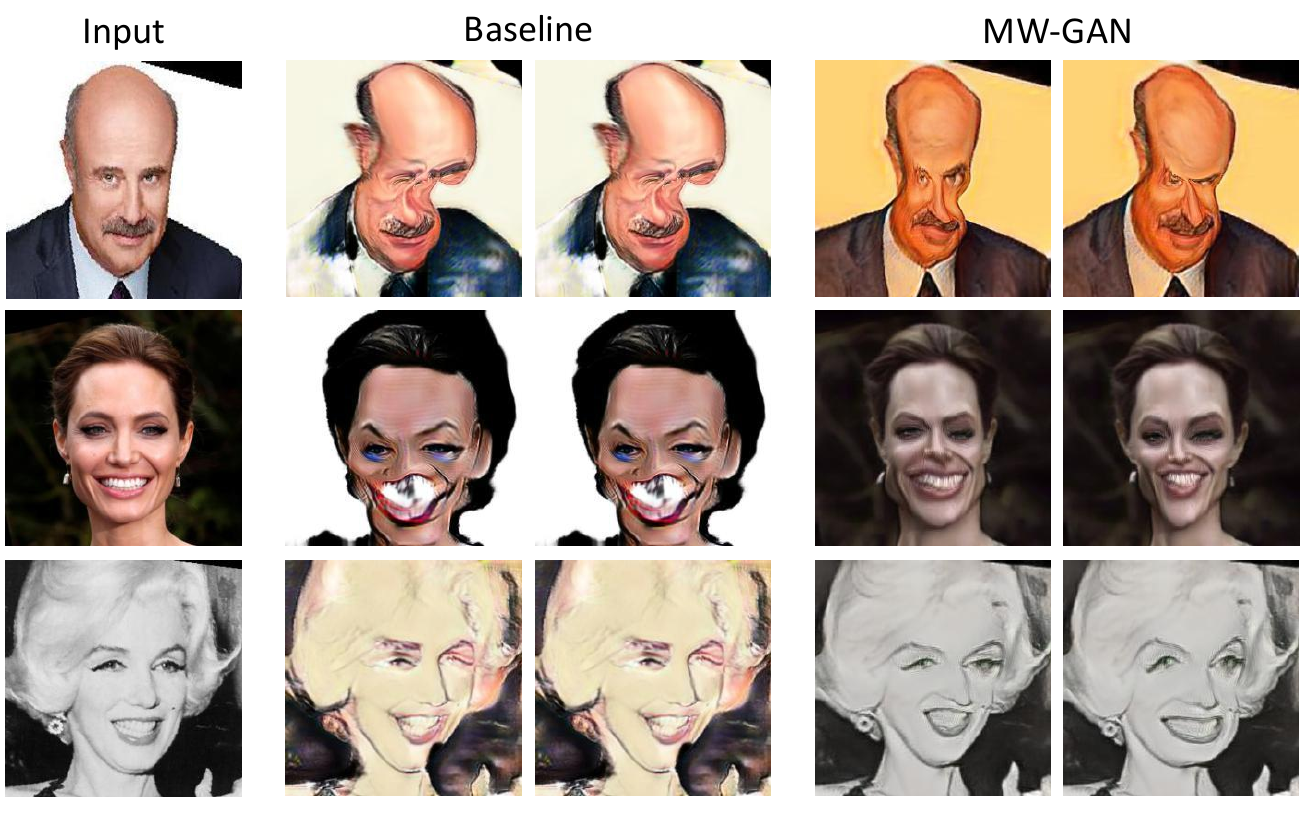}
  \caption{Comparison with the single way baseline method.}
  \label{ablation_dual}
\end{figure}

\subsection{Ablation Study}

To analyze our dual way design, geometric network training and identity recognition loss, we conducted experiments using the baseline method and three MW-GAN variants by respectively removing the GAN loss on generated landmarks ($L_{gan\_l}$), the identity recognition loss in the image space ($L_{id\_x}$) and landmark space ($L_{id\_l}$).
Other losses are either basic (reconstruction losses) or their effectiveness (GAN losses, cycle-consistency loss) have been demonstrated in other image translation or caricature generation methods~\cite{aae,cycleGAN,MUNIT,cariGANs,warpGAN}. They are therefore not included in this ablation study to avoid repetitive evaluations. \hou{For all these experiments, we have both qualitatively and quantitatively compared our MW-GAN with its three variants and the one-way baseline method. For quantitative evaluation, Fr{\'e}chet Inception Distance (FID)~\cite{fid} is used to evaluate the quality of the generated caricatures and rank-1 identification accuracy is used to evaluate the identity preservation ability. For the calculation of rank-1 accuracy, the ArcFace~\cite{arcface} model is adopted. The identification experiment was conducted where photos were kept in the gallery and the generated caricatures were used as probes. The rank-1 identification accuracy is then calculated accordingly.}



\begin{figure*}[t]
  \centering
  \includegraphics[width=0.98\textwidth]{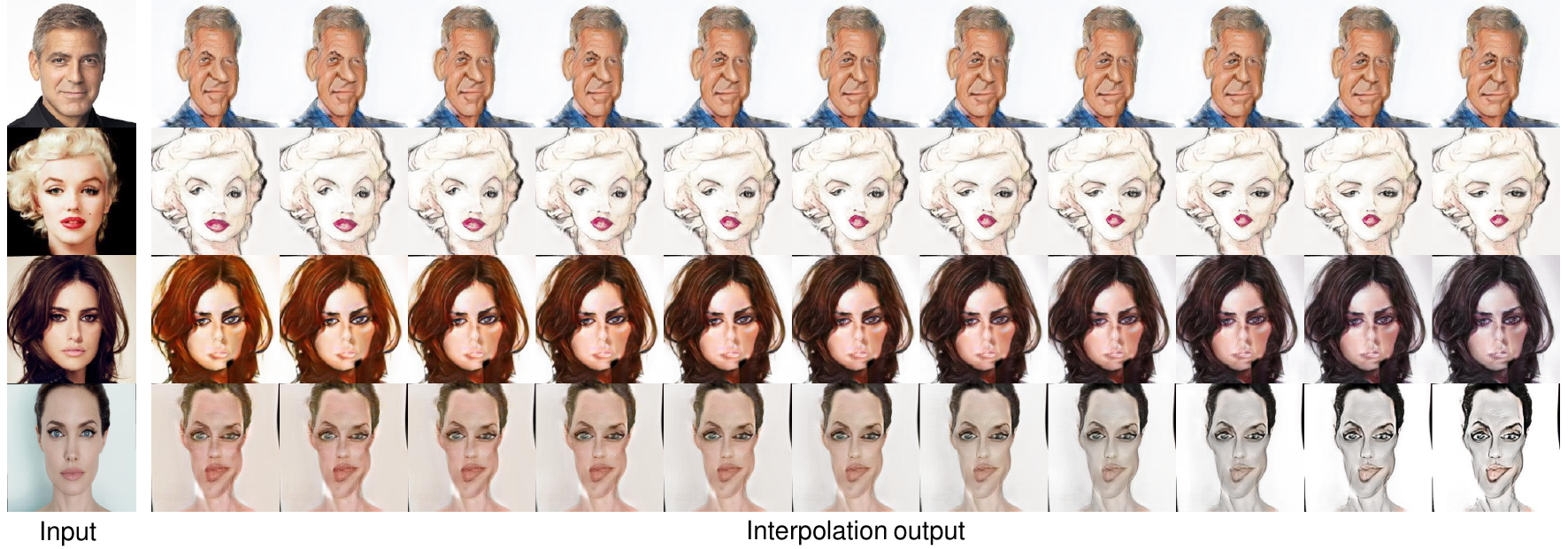}
  \caption{Interpolation experiment results. The top two rows are caricatures generated by interpolating the landmark transformation latent codes, while the bottom two rows are caricatures generated by interpolating the style latent codes.}
  \label{interpolation}
\end{figure*}

\begin{figure}[t]
  \centering
  \includegraphics[width=0.46\textwidth]{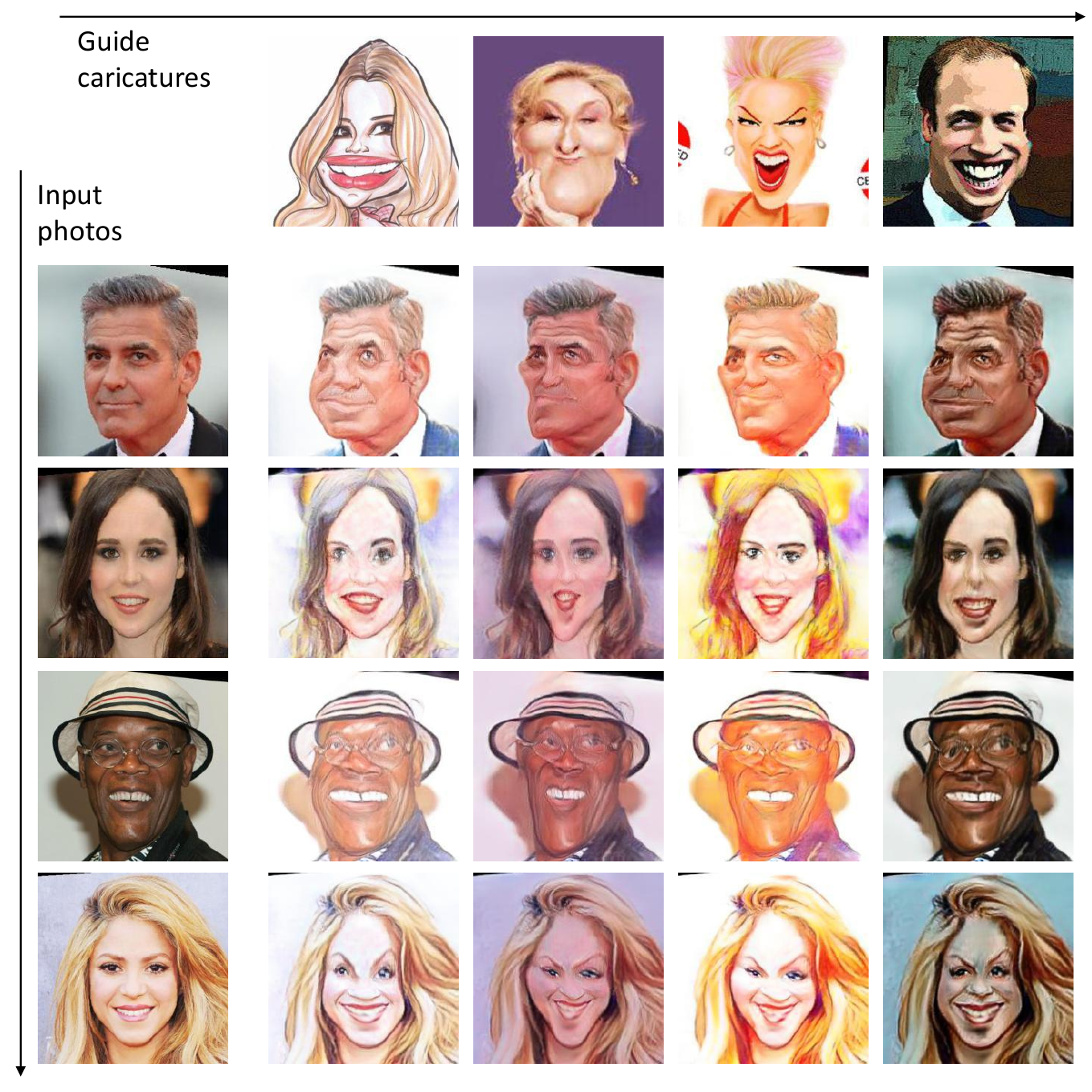}
  \caption{Sample-guided caricature generation of the proposed method.}
  \label{guide}
\end{figure}

\textbf{Study on different losses. }
Figure~\ref{ablation} shows the caricatures generated using MW-GAN and its three variants with different identity losses. From an overall view, it is obvious that caricatures generated by MW-GAN have much better visual quality than its other variants. At a closer look, the caricatures generated using the variant omitting the loss on images (\textit{w/o} $L_{id\_x}$) suffer from bad visual quality, as it lacks adversarial supervision from the image view. As for the variant omitting the loss on landmarks (\textit{w/o} $L_{id\_l}$), the generated caricatures have better visual quality, but their exaggerations are not in the direction to emphasize the subjects' characteristics. As for the variant omitting the GAN loss on generated landmarks (\textit{w/o} $L_{gan\_l}$), the exaggeration direction may break the facial components, since it lacks the supervision from adversarial learning, which helps with learning reasonable facial landmarks. By contrast, MW-GAN with both $L_{id\_x}$ and $L_{id\_l}$ can exaggerate facial shapes to enlarge the characteristics of the subjects, and meanwhile can render the caricatures with appealing texture styles. 

\hou{Additionally, as shown from the quantitative results in Table~\ref{ablation_table}, our GAN loss on generated facial landmarks, identity recognition loss in both landmark space and image space can greatly improve the quality of the generated caricatures (with lower FID scores). Besides, these losses also contribute to the identity preservation (with higher rank-1 accuracy). These observations are consistent with the qualitative analysis above.}

\textbf{Comparison with the single way baseline.} 
Figure~\ref{ablation_dual} shows the comparison of caricatures generated using our MW-GAN and using the baseline method described in Section~\ref{baseline_method}. For each input we randomly sample one style code and two landmark codes and generate two caricatures. From the results, we can see that with different landmark transformation codes, MW-GAN can generate caricatures with different exaggeration styles, while the single-way baseline method generates caricatures with almost the same exaggeration for each input. Moreover, it is also obvious that the exaggerations from the single-way method are sometimes out of control with unrealistic distortions, while our MW-GAN can generate much more meaningful exaggerations with the added cycle consistency supervision.

\hou{Besides, from the FID and rank-1 recognition accuracy of the baseline and MW-GAN in Table~\ref{ablation_table}, it is obvious that the quality and identity preservation ability of MW-GAN are much better than the baseline. This verifies our opinion in Section \ref{SubSec_MWGAN} that the dual way model can encode the content of the generated caricature backward and constrain it with the cycle loss. This makes the MW-GAN better in constraining the content of the generated caricature.}


\subsection{Diversity in Texture Style and Exaggeration}

In our MW-GAN network, the generated caricatures have their texture styles and shape exaggerations controlled by the style latent code and the landmark transformation latent code respectively. To achieve the diversity in generated caricatures, we can sample different style codes and landmark transformation codes, and apply them to the input photo. Figure~\ref{diversity} shows generated caricatures with different texture and exaggeration styles. For each input, we generate three caricatures with fixed exaggeration but different texture styles and another three caricatures with fixed texture style but different exaggerations. The results meet our expectation that different style codes lead to different texture and coloring, while different landmark codes lead to different shape exaggerations. 

Our dual-way design of MW-GAN enables unsupervised learning of the bidirectional mapping between image style and style latent space, geometric exaggeration and landmark transformation latent space. Therefore, MW-GAN can also generate caricatures with a guide sample by applying its style and landmark transformation codes to generators in the network. \hou{To do this, we firstly feed the guiding caricature into the style encoder and landmark encoder ($E_c^s$ and $E_c^l$) to get its style and landmark transformation codes. Then we use these codes as the style and landmark transformation codes in caricature generation.} Figure~\ref{guide} shows example caricatures generated with different guide samples. We can see that the generated caricatures not only have similar texture styles as the guide caricatures, but also try to mimic the exaggeration styles of the guide caricatures. From left to right, the generated caricatures have similar exaggeration styles of wide cheeks and narrow forehead, high cheekbones and squeezed facial features (eyes, nose and mouth), long face and pointed chin, and laughing with the mouth wide open.

\begin{figure*}[t]
  \centering
  \includegraphics[width=0.98\textwidth]{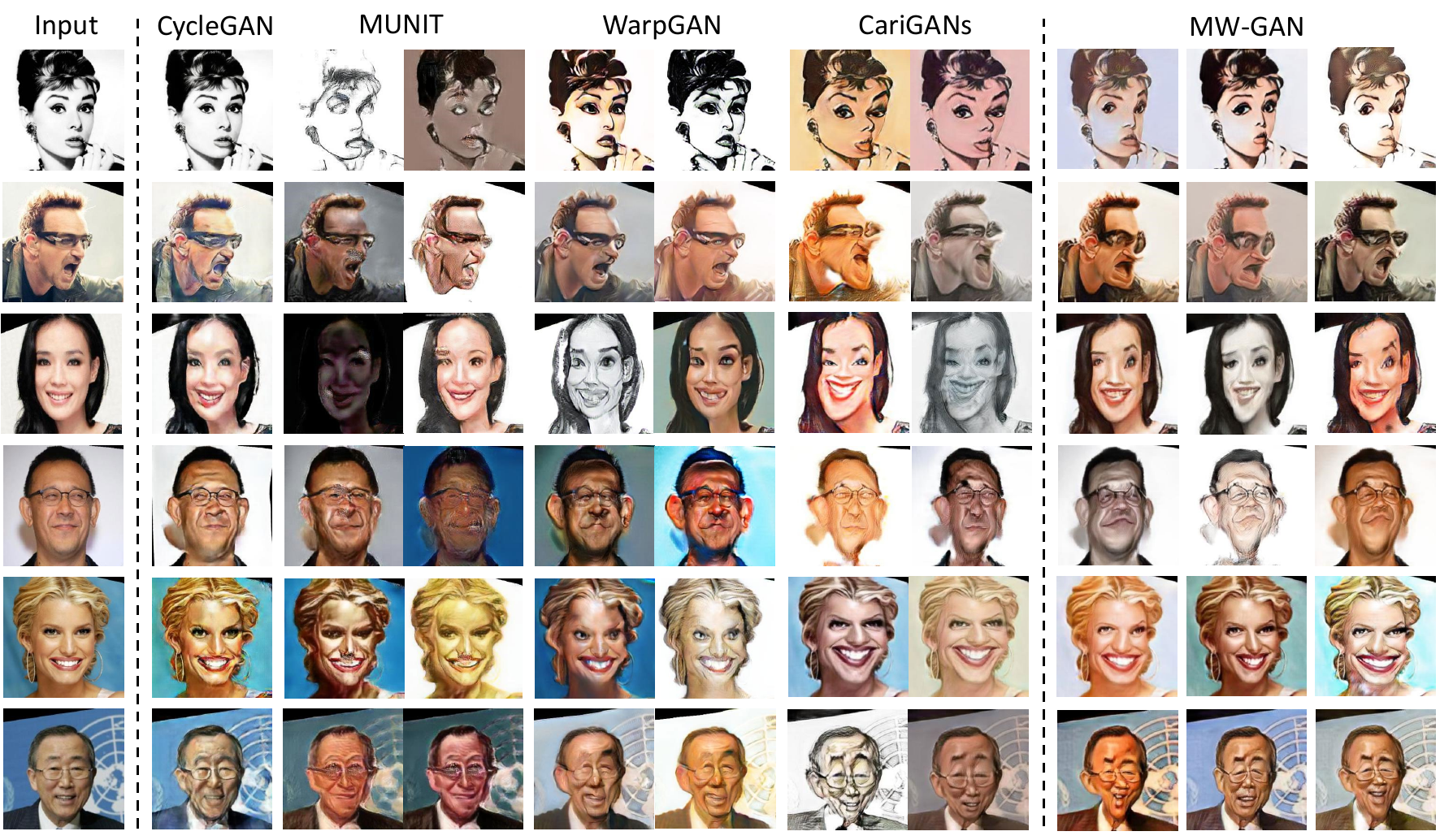}
  \caption{Comparison with state-of-the-art methods. CycleGAN and MUNIT can only generate images with style changed. Compared with WarpGAN and CariGANs, the proposed MW-GAN can generate caricatures with better visual quality and more flexible exaggeration. Besides, MW-GAN can generate caricatures with various styles and shape exaggerations.}
  \label{comparison}
\end{figure*}

\begin{table}
  \centering
  \caption{Comparison of FID and rank-1 accuracy (ACC) with state-of-the-art methods.}
  \label{comparison_table}
  \begin{tabular}{l c c c c c}
    \toprule
     & CycleGAN & MUNIT & WarpGAN & CariGANs & MW-GAN \\
    \midrule
    FID & 49.82 & 52.33 & 40.69 & 36.46 & \textbf{36.29}\\
    ACC & - & - & \textbf{78.55\%} & 57.95\% & 74.87\%\\
    \bottomrule
  \end{tabular}
\end{table}

\subsection{Interpolation Experiments}
As the texture style and exaggeration style are respectively controlled by style code and landmark transformation code in MW-GAN, 
we can achieve a `morphing' effect from one caricature to another by interpolating their codes (either color, exaggeration, or both).
For exaggeration interpolation, we randomly sample two landmark transformation latent codes ${z_c^l}_1$ and ${z_c^l}_2$, and generate caricatures with their interpolation $w{z_c^l}_1+(1-w){z_c^l}_2$, where $w$ ranges from 0 to 1 with step of 0.1. 
The texture style interpolation experiment is similarly conducted. Results are shown in Figure~\ref{interpolation}. 
We can see that the color and exaggeration style of caricatures change smoothly with different $w$. From left to right, 
the caricature face 
in the first row changes from thinner face with bigger nose to wider face with smaller nose. 
The face in the second row changes from bigger mouth to smaller ones. 
In the third row, the caricature color changes from orange face with red hair to purple face with black hair. 
In the last row, the caricature changes from a more colorful style to a grayer style. 
The smooth changes further demonstrate that the style and landmark latent codes are meaningfully learned in MW-GAN, and well represent the color and exaggeration styles of caricatures.

\subsection{Comparison with State-of-the-Art Methods}

We qualitatively and quantitatively compare our MW-GAN with previous state-of-the-art methods in image translation: CycleGAN~\cite{cycleGAN}, Multimodal UNsupervised Image-to-image Translation (MUNIT)~\cite{MUNIT}, and in caricature generation: WarpGAN~\cite{warpGAN}, CariGANs~\cite{cariGANs}. Since CariGANs are not open source and the data used in the paper is not publicly available, we implemented the CariGeoGAN using 17 landmarks with 34 dimension. The landmarks' dimension was reduced to 21, with 99.04\% of total variants preserved.

\begin{figure}[t]
  \centering
  \includegraphics[width=0.46\textwidth]{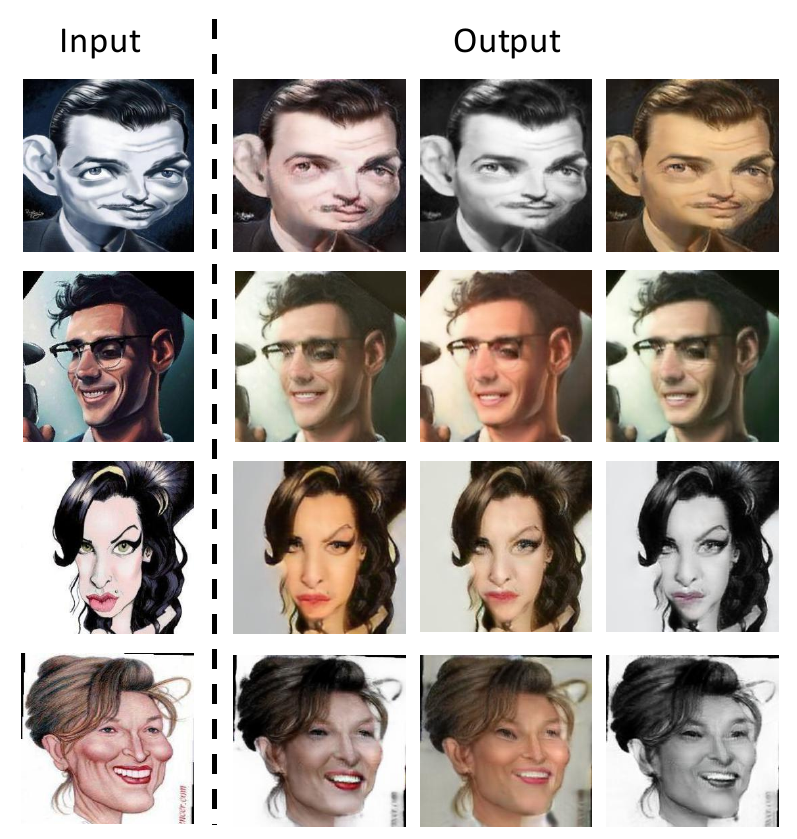}
  \caption{Translation of caricatures back to photos. The inputs are caricatures drawn by artists. The outputs are photos generated by MW-GAN with randomly sampled style latent codes and landmark latent codes.}
  \label{caricature2photo}
\end{figure}
 
Figure~\ref{comparison} shows the generated caricatures using different methods. When using MUNIT, WarpGAN and CariGANs, we randomly sample two style codes to generate two caricatures with different texture style for each input. When using our MW-GAN, we randomly sample style codes and landmark transformation codes and generate three caricatures for each input. CycleGAN is a deterministic method, and can only generate a fixed caricature for each input. As shown in Figure~\ref{comparison}, CycleGAN can only generate caricatures with limited changes in texture. Some results even look almost the same as the input. Caricatures generated by MUNIT have some changes in shape, but some results show clear artifacts, such as the speckles around the nose (2nd row) and the dark and patchy appearance (3rd row). As MUNIT is not designed for shape deformation, we speculate these artifacts arise from its attempt to achieve the appearance of shape deformation by some texture disguise. 
Results from WarpGAN and CariGANs can generate caricatures with more reasonable texture changes and shape exaggerations. However, given the input photo, their exaggeration style is deterministic, which does not reflect the diverse skills and preferences among artists. In comparison, our MW-GAN is designed to achieve diversity in both texture styles and shape exaggerations. As can be seen in Figure~\ref{comparison}, different style codes and landmark codes lead to different texture styles and shape exaggerations in the generated caricatures.

We also calculated FID to quantitatively measure the quality of generated caricatures (shown in Table~\ref{comparison_table}). 
Lower FID scores indicate the generated caricatures are more similar with real ones. Since CycleGAN and MUNIT are designed only for texture transformation but not geometric exaggeration as required in caricature generation, their FIDs are much higher than WarpGAN and MW-GAN, indicating lower quality. When it comes to WarpGAN, CariGANs and MW-GAN, it is shown that MW-GAN and CariGANs have a much lower FID than WarpGAN. 
MW-GAN and CariGANs have very similar FID scores, with
the FID of MW-GAN 
slightly lower. 
We believe this is due to three reasons. Firstly, MW-GAN specifically considers the exaggeration diversity, which is a closer assumption to the real distribution of caricatures. Secondly, the identity recognition loss in both image space and landmark space enables the learning of more meaningful shape exaggerations using MW-GAN. Finally, our dual-way design enables the learning of a bidirectional translation between caricatures and photos, and bridges the two with style latent codes and landmark latent codes. This design helps to train the model as a whole.

\hou{In order to quantify identity preservation accuracy for generated caricatures, we evaluated automatic face recognition performance for the three methods with geometric exaggeration (WarpGAN, CariGANs and our MW-GAN) using a state-of-the-art face recognition model, the ArcFace~\cite{arcface}. The identification experiment was conducted where photos were kept in the gallery while the generated caricatures were used as probes. We calculated the rank-1 identification accuracy and the results are shown in Table~\ref{comparison_table}. From the results, we can see that CariGANs has the lowest identity preservation accuracy, while WarpGAN and MW-GAN both have accuracy over 70\%. Though the identity preservation accuracy of MW-GAN is a little lower than that of WarpGAN, we consider 
that exaggerations without diversity are easier to achieve a high identity preservation accuracy, and the following perceptual study also confirms that the diverse exaggerations generated by MW-GAN are reasonable.}

\subsection{Perceptual Study}
We conducted a perceptual study to evaluate the generated caricatures in terms of their 1) visual quality, i.e., whether the generated caricatures are visually appealing; and 2) exaggeration quality, i.e., whether the exaggerated face is recognizable.
For each photo in the test set, $5$ caricatures were generated using the five methods: CycleGAN, MUNIT, Warp-GAN, CariGANs, and our MW-GAN. 
In each study of visual quality and exaggeration quality, volunteers are shown an input photo and the corresponding caricatures generated by the five methods and were asked to vote for the best and worst ones from the five caricatures. We randomly selected $100$ test photos and their generated caricatures, and presented them to $13$ volunteers for voting. That is $1300$ votes in total. One best vote counts +1, while one worst vote counts -1. The final score for each method is the average of counts. Results are shown in Table~\ref{perceptual_study}.

It is obvious that WarpGAN, CariGANs and MW-GAN have much better exaggeration quality than CycleGAN and MUNIT, as these caricature generation methods explicitly exaggerate geometric shapes through warping. Among the three, CariGANs and MW-GAN are the best with comparable exaggeration quality. However, MW-GAN can generate caricatures with various exaggeration styles from the same input photo, while CariGANs can only exaggerate the face in a certain way.
When it comes to the visual quality, we can see that MW-GAN has a higher score than others. We reason that it is because the explicit encoding of the variety of color and exaggeration styles better captures the distribution of real caricatures. MUNIT has the worst visual quality, as the content code reconstruction in MUNIT is contradictory to the fact that caricatures have different shape structures from photos. In summary, the exaggerations generated by MW-GAN are both reasonable and diverse, and the caricatures generated by MW-GAN are the most visually appealing among the five methods, according to the perceptual study.

\begin{table}
  \centering
  \caption{Perceptual study on exaggeration quality (EQ) and visual quality (VQ).}
  \label{perceptual_study}
  \begin{tabular}{l c c c c c}
    \toprule
    Method & CycleGAN & MUNIT & WarpGAN & CariGANs & MW-GAN \\
    \midrule
    EQ-best & 71 & 21 & 290 & \textbf{464} & 454\\
    EQ-worst & 625 & 407 & 91 & 91 & \textbf{86}\\
    EQ-score & -0.426 & -0.297 & 0.153 & \textbf{0.287} & 0.283\\
    VQ-best & 176 & 9 & 305 & 396 & \textbf{414}\\
    VQ-worst & \textbf{40} & 956 & 113 & 122 & 69\\
    VQ-score & 0.105 & -0.728 & 0.148 & 0.211 & \textbf{0.265}\\
    \bottomrule
  \end{tabular}
\end{table}

\subsection{Translation of Caricatures Back to Photos}
Although not our main goal, 
there is also a path in our network architecture to translate caricatures back to photos because of the dual way design. We conducted experiments by using caricatures and their landmarks as input, and randomly 
sampled
style latent code $z_p^s$ and landmark latent code $z_{p}^l$ for photos from the corresponding distributions. The generated photos are shown in Figure~\ref{caricature2photo}. The results show that MW-GAN can restore photos from caricatures by reversely deforming the exaggerated faces back to normal. As can be seen in the 2nd and 4th rows, the shapes of the cheek, chin and mouth in the generated photos look realistic. It shows the capability of MW-GAN to moderately deform the caricature shapes to photo shapes. However, as deformations from caricatures to photos are more complicated, the 17 landmarks are often insufficient to represent such deformations. When the input caricature has extreme exaggeration, MW-GAN may generate the photo without sufficient 
shape deformation, resulting residual deformations from the input caricature e.g., the ear and the chin in the 1st row of Figure 11, and the eyes and cheek in the 3rd row. of Figure~\ref{caricature2photo}.
Further exploration is required for translations of caricatures back to photos.

\subsection{Limitations and Future Work}
As the first framework to generate caricatures with diversities in both texture styles and shape exaggerations, MW-GAN still has some limitations to 
be improved in the future. Firstly, 
when translating caricatures back to photos, because of the more complicated deformation, the results are not desirable.
To address this problem, 
further explorations are required. One possible direction is to use more flexible deformation representation.
Secondly, not all of the generated exaggerations are plausible. For example, some generated caricatures have eyes with different sizes e.g., the 3rd row in Figure~\ref{comparison}, which may be caused by the asymmetry of the hairs around the eyes. Therefore, further improvements of the framework and loss design could be explored to generate caricatures with higher quality. \hou{
MW-GAN is proposed as a base framework to generate caricatures with diverse style and exaggerations. More explorations, such as using better GAN network and other improvement to this framework, are also worth further study.} 
\hou{Lastly, as evaluation of the exaggeration quality is subjective which varies from person to person, we consider another potential direction is to combine user interaction with the proposed method to allow further adjustment of the exaggeration results.}

\section{Conclusion}
In this paper, we propose the first framework that can generate caricatures with diversities in both texture styles and shape exaggerations. In our design, we use style latent code and landmark transformation latent code to capture the diversity in texture and exaggeration respectively. We also design a dual way framework to learn the bidirectional translation between photos and caricatures. This design helps the model to learn the bidirectional translation between the image style, face landmarks and their corresponding latent spaces, which enables the generation of caricatures with sample-guided texture and exaggeration styles. We also introduced identity recognition loss in both image space and landmark space, which enables the model to learn more meaningful exaggeration and texture styles for the input photo. Qualitative and quantitative results demonstrate that MW-GAN outperforms the state-of-the-art methods in image translation and caricature generation. Future work includes improving the geometric exaggeration quality, incorporating user interactions and exploring better GAN network structures. The problem of translating caricatures to photos is also worth studying.



%





\ifCLASSOPTIONcaptionsoff
  \newpage
\fi



\bibliographystyle{IEEEtran}
\bibliography{IEEEabrv,mwgan_bib}
%



%

\begin{IEEEbiography}[{\includegraphics[width=1in,height=1.25 in,clip,keepaspectratio]{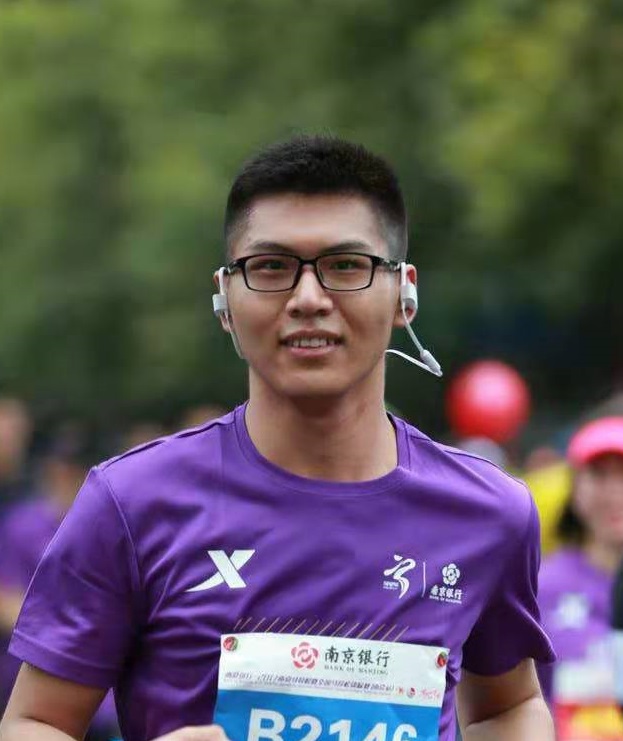}}]{Haodi Hou}
received the BSc from Nanjing University, China in 2017. Currently, he is a MSc candidate in the Department of Computer Science and Technology at Nanjing University and a member of RL Group, which is led by professor Yang Gao. His research interests are mainly on Neural Networks, Generative Adversarial Nets and their application in image translation and face recognition.
\end{IEEEbiography}

\begin{IEEEbiography}[{\includegraphics[width=1in,height=1.25in,clip,keepaspectratio]{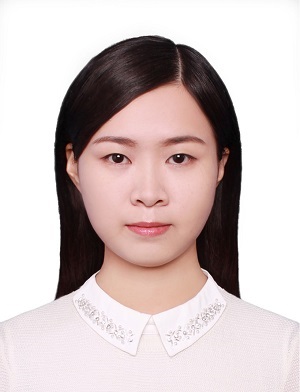}}]{Jing Huo} received the PhD degree from the Department of Computer Science and Technology, Nanjing University, Nanjing, China, in 2017. She is currently an Assistant Researcher with the Department of Computer Science and Technology, Nanjing University. Her current research interests include machine learning and computer vision, with a focus on subspace learning, adversarial learning and their applications to heterogeneous face recognition and cross-modal face generation. 
\end{IEEEbiography}

\begin{IEEEbiography}[{\includegraphics[width=1in,height=1.25in,clip,keepaspectratio]{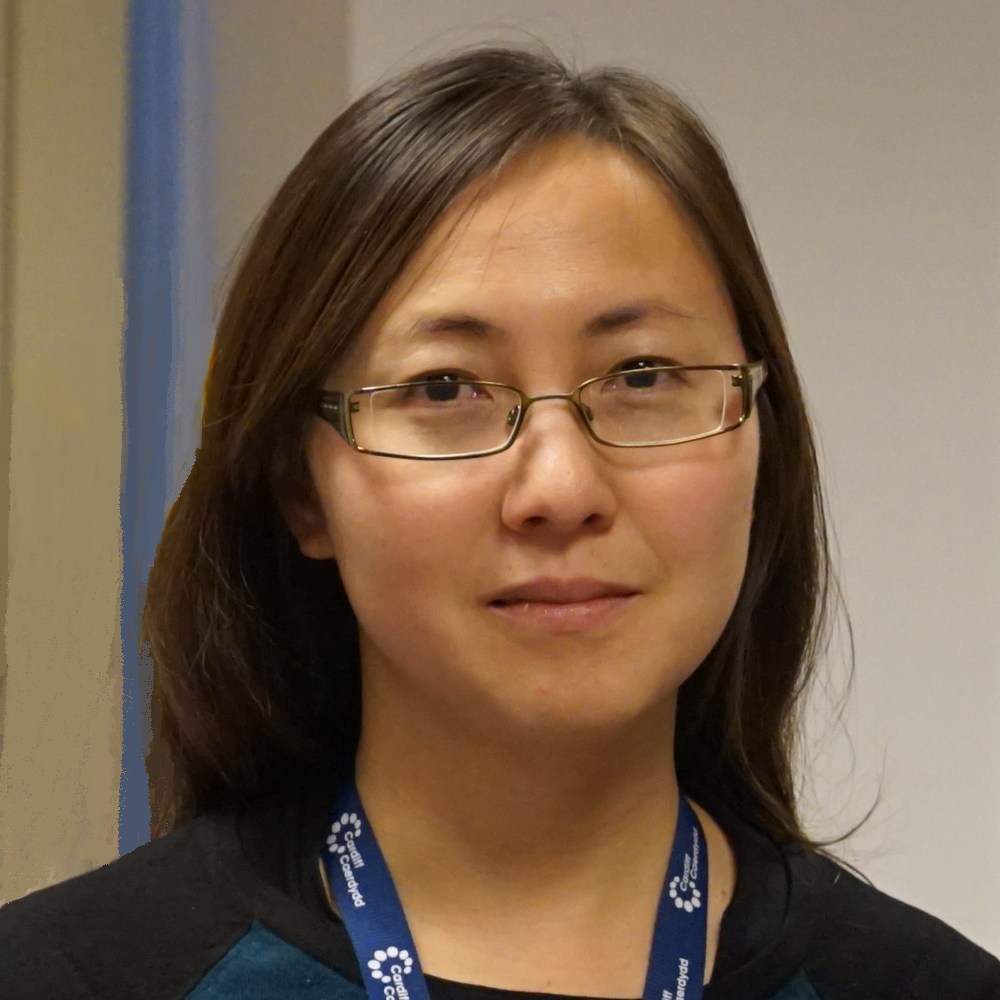}}]{Jing Wu}
received the BSc and MSc degrees from Nanjing University, China in 2002 and 2005 respectively. She received the PhD degree in computer science from the University of York, UK in 2009. She worked as a research associate in Cardiff University subsequently, and is currently a lecturer in computer science and informatics at Cardiff University. Her research interests are on image-based 3D reconstruction and its applications. She is a member of ACM and BMVA.
\end{IEEEbiography}

\begin{IEEEbiography}[{\includegraphics[width=1in,height=1.25in,clip,keepaspectratio]{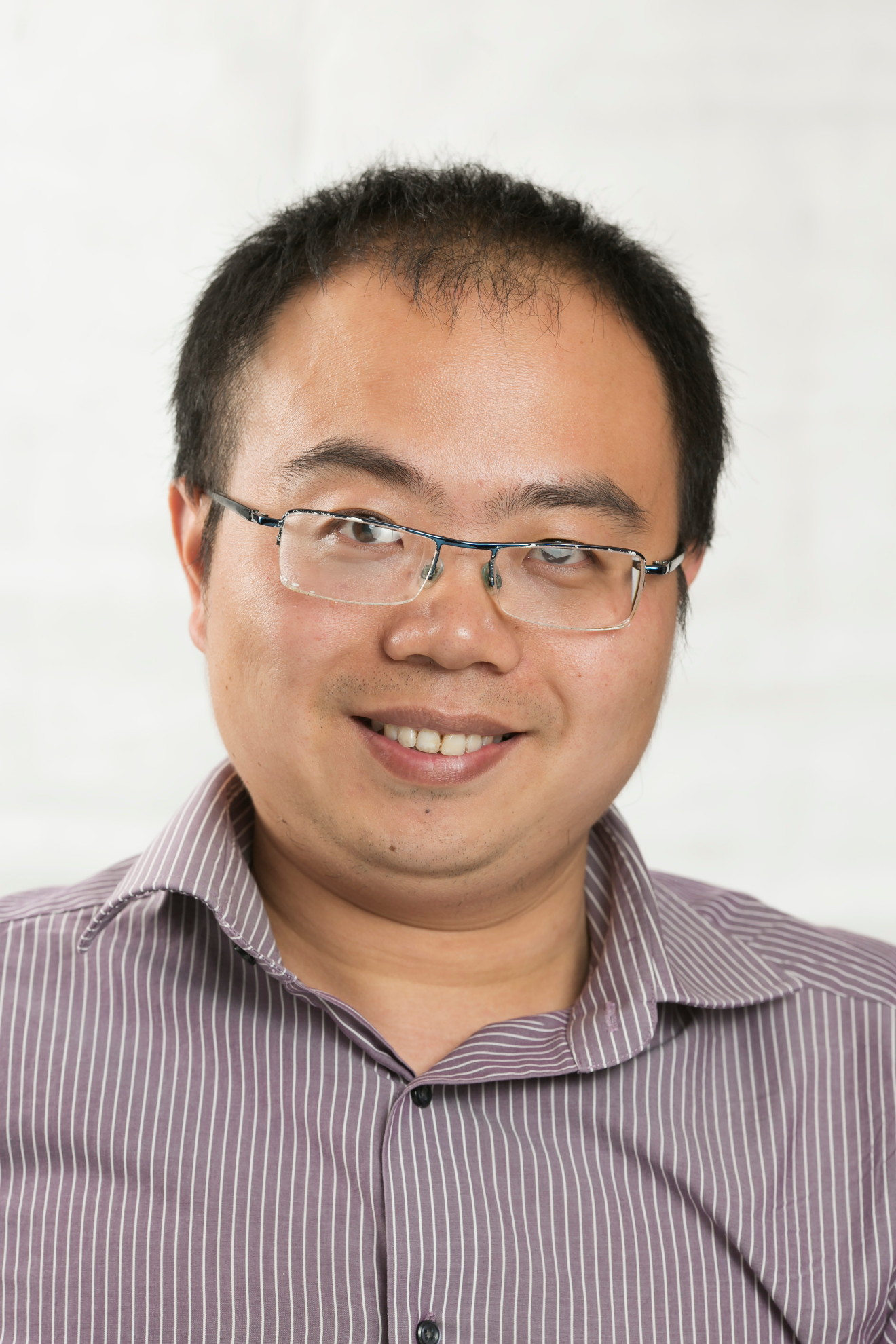}}]{Yu-Kun Lai} received the bachelor’s and PhD degrees in computer science from Tsinghua University, in 2003 and 2008, respectively. He is currently a Professor with the School of Computer
Science \& Informatics, Cardiff University. His 
research interests include computer graphics, 
geometry processing, image processing, and 
computer vision. He is on the editorial boards of
\emph{Computer Graphics Forum} and \emph{The Visual Computer}.

\end{IEEEbiography}

\vfill

\begin{IEEEbiography}[{\includegraphics[width=1in,height=1.25in,clip,keepaspectratio]{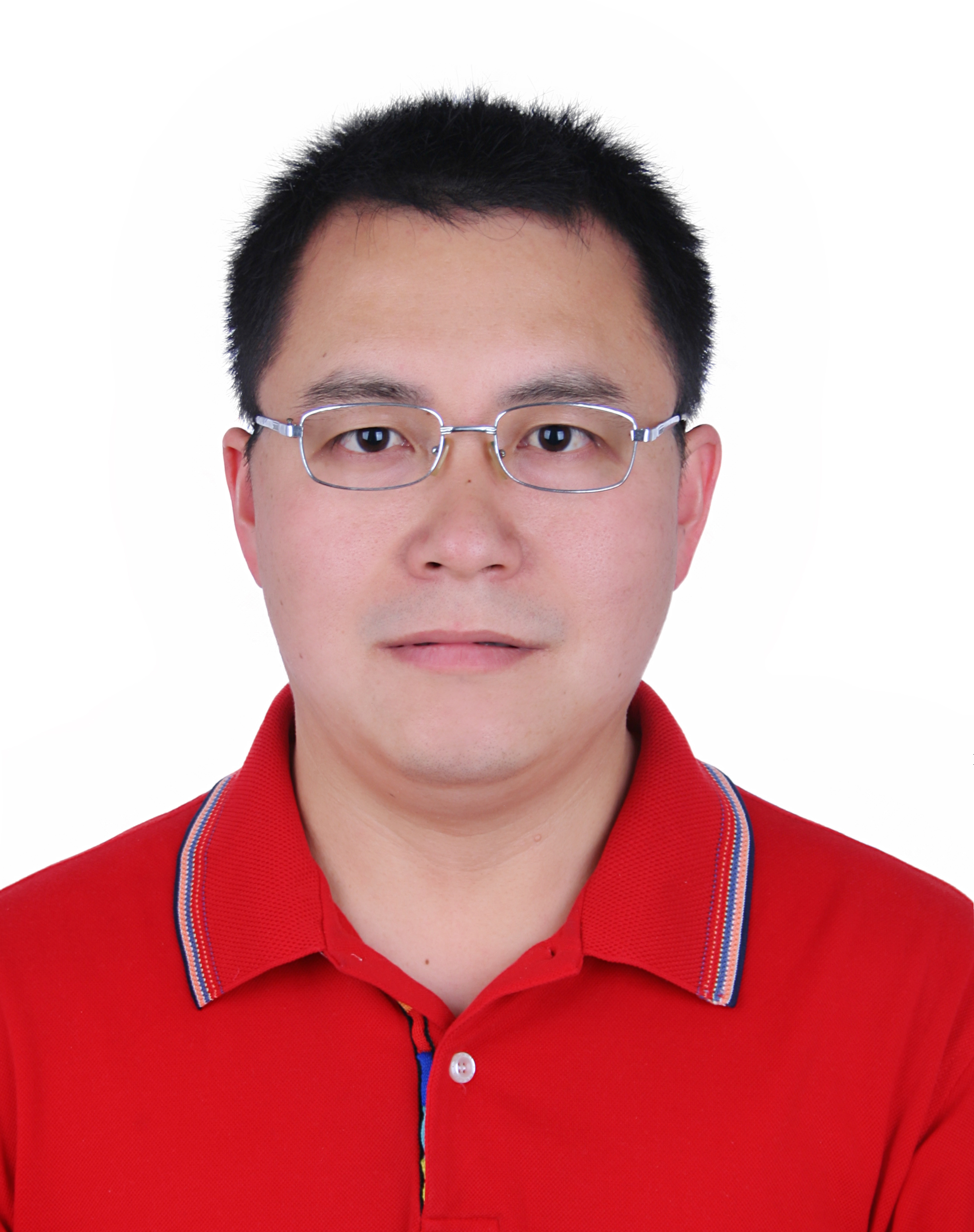}}]{Yang Gao} (M’05) received the Ph.D. degree in computer software and theory from Nanjing University, Nanjing, China, in 2000. He is a Professor with the Department of Computer Science and Technology, Nanjing University. He has published over 100 papers in top conferences and journals. His current research interests include artiﬁcial intelligence and machine learning.

\end{IEEEbiography}

\vfill
\nopagebreak[4]









\end{document}